\documentclass[a4paper]{article}

\usepackage{INTERSPEECH2019}
\usepackage{multirow}
\usepackage{booktabs}
\usepackage{algorithm}  
\usepackage{algpseudocode}  
\usepackage{amsmath}  
\title{Low-bit Quantization of Recurrent Neural Network Language Models Using Alternating Direction Methods of Multipliers}
\name{Junhao Xu$^\star$, Xie Chen$^\dagger$, Shoukang Hu$^\star$ ,Jianwei Yu$^\star$,  Xunying Liu$^\star$, Helen Meng$^\star$}
\address{
  $^\star$The Chinese University of Hong Kong, Hong Kong SAR, China\\
  $^\dagger$Microsoft AI and Research, One Microsoft Way, Redmond, WA, USA}
\email{\{jhxu, skhu, jwyu, xyliu, hmmeng\} @se.cuhk.edu.hk,  xieche@microsoft.com}

\begin{document}

\maketitle
\begin{abstract}
  The high memory consumption and computational costs of Recurrent neural network language models (RNNLMs) limit their wider application on resource constrained devices. In recent years, neural network quantization techniques that are capable of producing extremely low-bit compression, for example, binarized RNNLMs,  are gaining increasing research interests. Directly training of quantized neural networks is difficult. By formulating quantized RNNLMs training as an optimization problem, this paper presents a novel method to train quantized RNNLMs from scratch using alternating direction methods of multipliers (ADMM). This method can also flexibly adjust the trade-off between the compression rate and model performance using tied low-bit quantization tables. Experiments on two tasks: Penn Treebank (PTB), and Switchboard (SWBD) suggest the proposed ADMM quantization achieved a model size compression factor of up to 31 times over the full precision baseline RNNLMs. Faster convergence of 5 times in model training over the baseline binarized RNNLM quantization was also obtained.
  
\end{abstract}

\section{Introduction}
RNNLMs are widely used in state-of-art speech recognition systems. The high memory consumption and computational costs limit their wider application on resource constrained devices. In order to address this issue for RNNLMs, and deep learning in general, a wide range of deep model compression appproaches including teacher-student based transfer learning~\cite{hinton2015distilling}~\cite{chebotar2016distilling}~\cite{huang2018knowledge}, low rank matrix factorization~\cite{sainath2013low}~\cite{jaderberg2014speeding}~\cite{lebedev2014speeding}~\cite{tai2015convolutional}~\cite{sindhwani2015structured}, sparse weight matrices~\cite{liu2015sparse}~\cite{han2015learning}\cite{han2015learning}~\cite{wen2016learning} have been proposed. In addition, a highly efficient family of compression techniques based on deep neural network quantization that are capable of producing extremely low bit representation~\cite{courbariaux2014training}~\cite{courbariaux2015binaryconnect}~\cite{courbariaux2016binarized}, for example, binarized RNNLMs~\cite{liu2018binarized},  are gaining increasing research interests. 

Earlier forms of deep neural network (DNN) quantization methods compress well-trained full precision models off-line~\cite{gong2014compressing}~\cite{chen2015compressing}. In~\cite{chen2015compressing}, a hash function was used to randomly group the connection weight parameters into several shared values. Weights in convolutional layers are quantized in~\cite{gong2014compressing}. In order to reduce the inconsistency in error cost function between the full precision model training and subsequent quantization stages, later researches aimed at directly training a low bit neural network from scratch~\cite{soudry2014expectation}~\cite{courbariaux2016binarized}~\cite{liu2018binarized}.

The key challenge in these approaches is that gradient descent methods and back-propagation (BP) algorithm can not be directly applied in quantized model training when the weights are restricted to discrete values. To this end, there have been two solutions to this problem in the machine leaning community~\cite{soudry2014expectation}~\cite{courbariaux2016binarized}. A Bayesian approach was proposed in~\cite{soudry2014expectation} to allow a posterior distribution over discrete weight parameters to be estimated, and the subsequent model quantization used samples drawn from the distribution. In~\cite{courbariaux2016binarized}, low precision binarized parameters were first used in the forward pass to compute the error loss before full precision parameters  are used in the backward pass to propagate the gradients.  It was further suggested in~\cite{liu2018binarized} for RNNLMs that extra partially quantized linear layers containing binary weight matrices, full  precision bias and additional scaling parameters need to be added to mitigate the performance degradation due to model compression. A compression ratio of 11.3 in~\cite{liu2018binarized} was reported on PTB and SWBD data without performance loss.

In this paper, by formulating quantized RNNLMs training as an optimization problem, a novel method based on alternating direction methods of multipliers (ADMM)~\cite{boyd2011distributed}~\cite{leng2018extremely} is proposed. Two sets of parameters: a full precision network, and the optimal quantization table, are considered in a decomposed dual ascent scheme and optimized in an alternating fashion iteratively using an augmented Lagrangian. This algorithm draws strength from both the decomposibility of the dual ascent schemes and the stable convergence of multiplier methods. In order to account for the detailed parameter distributions at a layer or node level within the network, locally shared quantization tables trained using ADMM are also proposed to allow fine-grained and flexible adjustment over the trade-off between model compression rate and performance. 

The main contributions of this paper are summarized as below. First, to the best of our knowledge, this paper is the first work to introduce ADMM based RNNLMs quantization for speech recognition tasks. The previous research on low bit quantization~\cite{courbariaux2016binarized} of RNNLMs~\cite{liu2018binarized} focused on using different parameter precision in the error forwarding and gradient backward propagation stages. 
%% The use of quantization was also restricted to certain parts of the network only, for example, the embedding layers. In contrast, our methods allow low bit quantization to be applied to all layers. 
The earlier use of alternating methods in off-line quantization of well trained DNNs~\cite{xu2018alternating} also does not allow low bit quantized models to be directly trained from scratch, as considered in this paper. Second, the previous application of ADMM DNN quantization was restricted to computer vision tasks~\cite{leng2018extremely}. In addition, a globally tied quantization table was applied to all parameters in the network, thus providing limited flexibility to account for detailed local parameter distributions as considered in this paper. We evaluate the performance of the proposed ADMM RNNLMs quantization method on two tasks targeting primarily on speech recognition applications: Penn Treebank and Switchboard, in comparison against the baseline binarized RNNLMs quantization in terms of the trade-off between model compression factor, perplexity and speech recognition error rate.

The rest of the paper is organized as follows. RNNLMs are reviewed in section 2. 
A general neural network quantization scheme is described in section 3. Section 4 presents our ADMM based RNNLMs quantization in detail. Experiments and results are shown in section 5. Finally, conclusions and future work are discussed in section 6.

\section{Recurrent Neural Network LMs}
\label{sec:3}
The recurrent neural network language models (RNNLMs) we considerred in this paper computes the word probability by

 \begin{align}
  \vspace{-1cm} 
      P(\mathbf{w}_{t}|\mathbf{w}^{t-1}_{1})\approx P\left(\mathbf{w}_{t}|\mathbf{w}_{t-1}, \mathbf{h}_{t-1}\right),
  \vspace{-1cm} 
  \end{align}
  
 where $\mathbf{h}_{t-1}$ is the \textit{hidden state} that attempts to encode the history information $(\mathbf{w}_{t-2},...,\mathbf{w}_{1})$ into a $D$-dimensional vector representation, where $D$ is the number of hidden nodes.

  In RNNLMs, a word $\mathbf{w}_{t}$ is represented by a $N$-dimensional one-hot vector $\mathbf{\tilde{w}}_{t}$, where $N$ is the vocabulary size. To process sparse data, the one-hot vector is first projected into a $M$-dimensional size ($M\ll N$) continuous space \cite{bengio2003neural} where $M$ is considered as he embedding size:
  \begin{align}
    \vspace{-1cm} 
    \mathbf{x}_{t}=\mathbf{\Theta}_U\mathbf{\tilde{w}}_{t}^\top,
    \vspace{-1cm} 
    \end{align}

where $\mathbf{\Theta}_U$ is a projection matrix to be trained. After the word embedding layer, the hidden state $\mathbf{h}_{t}$ is calculated recursively through a gating function:
$\mathbf{h}_{t}=\mathbf{g}(\mathbf{h}_{t-1}, \mathbf{x}_{t-1})$,
which is a vector function that controls the amount of inherited information from $\mathbf{w}_{t-1}$ in the current ``memory'' state $\mathbf{h}_{t}$. 
Currently, long short-term memory (LSTM) \cite{hochreiter1997lstm} RNNLMs\cite{sundermeyer2012lstm} definite the state of art performance.

\par
In order to solve the problem of vanishing gradients, LSTM introduces another recursively computed variable $\mathbf{c}_t$, a \textit{memory cell}, which aims to preserve the historical information over a longer time window. At time $t$ four gates are computed -- the forget gate $\mathbf{f}_t$, the input gate $\mathbf{i}_t$, the cell gate $\mathbf{\tilde{c}}_t$ and the output gate $\mathbf{o}_t$:
\begin{align}
\vspace{-1cm} 
    \label{eq:f-gate}
    \mathbf{f}_t=&\, \boldsymbol{\sigma}\left(\boldsymbol{\Theta}_\text{f}\left[\mathbf{x}_{t-1}, \mathbf{h}_{t-1}, 1\right]^\top\right)\\
    \label{eq:i-gate}
    \mathbf{i}_t=&\, \boldsymbol{\sigma}\left(\boldsymbol{\Theta}_\text{i}\left[\mathbf{x}_{t-1}, \mathbf{h}_{t-1}, 1\right]^\top\right)\\
    \label{eq:c-gate}
    \mathbf{\tilde{c}}_t=&\, \tanh\left(\boldsymbol{\Theta}_\text{c}\left[\mathbf{x}_{t-1}, \mathbf{h}_{t-1}, 1\right]^\top\right)\\
    \label{eq:o-gate}
    \mathbf{o}_t=&\, \boldsymbol{\sigma}\left(\boldsymbol{\Theta}_\text{o}\left[\mathbf{x}_{t-1}, \mathbf{h}_{t-1}, 1\right]^\top\right)
\vspace{-1cm} 
\end{align}
where $\tanh(\mathbf{u})=[\tanh(u_1), ..., \tanh(u_D)]$ for any $\mathbf{u}\in\mathbb{R}^D$. With the four gating outputs, we update
\begin{align}
    \label{eq:c_t}
    \mathbf{c}_t=&\,\mathbf{f}_t\otimes \mathbf{c}_{t-1} + \mathbf{i}_t\otimes \mathbf{\tilde{c}}_t\\
    \label{eq:h_t}
    \mathbf{h}_{t}=&\,\mathbf{o}_t\otimes\tanh(\mathbf{c}_t),
\vspace{-1cm} 
\end{align}
where $\otimes$ is the Hadamard product. 

\section{Neural Network Quantization}

The standard n-bit quantization problem for
neural network considers for any full precision weight parameter, $\theta$, finding its closest discrete approximation from 
the following the quantization table.

\begin{equation}
\label{equ:simQuan}
q \in \{0, \pm1,\pm2, \dots, \pm 2^n\}
\end{equation}

as 

\begin{equation}
  \begin{split}
  \label{eq:quanMap}
  f(\theta) = \arg\min \limits_{q} |\theta - q|
\vspace{-1cm} 
\end{split}
\end{equation}

Further simplification to the above quantization table of Equation (\ref{equ:simQuan}) leads to either the binarized $\{-1, 1\}$~\cite{rastegari2016xnor}, or tertiary value  $\{-1, 0, 1\}$~\cite{li2016ternary} based quantization. 

It is assumed in the above quantization that a global quantization table is applied to all weight parameters. 
In order to account for the detailed local parameter distributional properties, and more importantly flexibly adjust the trade off between model compression ratio and performance degradation, the following more general form of quantization is considered for each parameter $\Theta^{(l)}_{i}$ within any of the $l^{th}$ weight cluster, for example, all weight parameters of the same layer,

 \begin{equation}
  \begin{split}
  \label{eq:quanMap}
  f(\Theta^{(l)}_i) = \arg\min \limits_{Q_i^{(l)}} |\Theta^{(l)}_i - Q_i^{(l)}|
\vspace{-1cm} 
\end{split}
\end{equation}
can be used. The locally shared $l^{th}$ quantization table is given by
\begin{equation}
\label{equ:31}
  {Q}_i^{(l)}\in\{0, \pm\alpha^{(l)},\dots, \pm\alpha^{(l)} 2^n\}
\end{equation}

$\alpha^{(l)}$ is used to represent the scaling factor of the original discrete quantization.  It is shared locally among weight parameters clusters. The tying of quantization tables may be flexibly performed at either node, layer level, or in the extreme case individual parameter level (equivalent to no quantization being applied). 

Intuitively, the larger quantization table is used, a smaller compression rate after quantization and reduced performance degradation is expected.   A projection from the original full precision parameters to the quantized low precision values needs to be found using Equation (\ref{eq:quanMap}) during the training phase.

\section{RNNLMs Quantization Using ADMM}
\label{sec:4}
Alternating direction methods of multipliers (ADMM) is a powerful optimization technique. It decomposes a dual ascent problem into alternating updates of two variables. In the context of the RNNLMs quantization problem considered here, these refer to full precision model weights update and the discrete quantization table estimation.  In addition to the standard Lagrangian term taking the form of a dot production between the multiplier variable $\boldsymbol\lambda$ and the quantization error $(\boldsymbol{\Theta}-f(\boldsymbol{\Theta}))$, it is also useful for alternating direction methods to introduce an additional term to form an augmented Lagrangian~\cite{boyd2011distributed} to improve robustness and convergence speed of the algorithm. The overall Lagrange function is formulated as:

\begin{align}
\label{equ:AL}
  L = \mathcal{F}_{ce}(\boldsymbol{\Theta})+(\gamma \boldsymbol\lambda)^\top\cdot(\boldsymbol{\Theta}-f(\boldsymbol\Theta)) + \frac{\gamma}{2}||\boldsymbol{\Theta}-f(\boldsymbol\Theta)||_2^2
  \end{align}

where $\mathcal{F}_{ce}$ is the crossentropy loss of the neural network, $\boldsymbol{\Theta}$ are the network parameters. $f(\boldsymbol\Theta)$ represents the quantization of the parameters calculated from the projection (\ref{eq:quanMap}). $\gamma>0$ is the penalty parameter and $\boldsymbol\lambda$ denotes the Lagrangian multiplier. 

Further rearranging Equation (\ref{equ:AL})  leads to the following loss function.
%%Then, after several mathematical transformations, we can easily reformulated the Lagrange function as following:
\begin{align}
\label{equ:AL2}
  L = \mathcal{F}_{ce}(\boldsymbol{\Theta}) +  \frac{\gamma}{2}||\boldsymbol{\Theta}-f(\boldsymbol\Theta)+\boldsymbol{\lambda}||_2^2-\frac{\gamma}{2}||\boldsymbol{\lambda}||^2
  \end{align}
  
  The algorithm when being performed at the $(k+1)^{th}$ iteration includes three stages.
  For simplicity, we assume a globally shared quantization table $\{0, \pm\alpha,\dots, \pm\alpha 2^n\}$ with a single scaling factor $\alpha$ to be learned. The following iterative update can be  extended when multiple shared quantization tables and associated scaling factors in Equation~(\ref{equ:31}) are used.
 
 \vspace{0.5em}
\noindent\textbf{1. Full precision weight update}
 \vspace{0.5em}
 
 \noindent The following equation is used to update the full precision weight parameters $\boldsymbol\Theta^{(k+1)}$.
 
%% From standard ADMM training procedure, we use the following equation to update weights parameters:
\begin{equation}
  \boldsymbol{\Theta}^{(k+1)} = \arg\min \limits_{\boldsymbol{\Theta}} L(\boldsymbol{\Theta}, f(\boldsymbol\Theta^{(k)}),\boldsymbol\lambda^{(k)})
\end{equation}
where $f(\boldsymbol\Theta^{(k)}), \lambda^{(k)}$ are the quantized weights and error variable at the $k^{th}$ iteration. The gradient of the loss function in Equation (\ref{equ:AL2}) w.r.t  $\boldsymbol\Theta$ is calculated as the following.

\begin{equation}\label{equ:gradient}
  \nabla L=\nabla\mathcal{F}_{ce}+\gamma(\boldsymbol{\Theta} - f(\boldsymbol\Theta^{(k)})+\boldsymbol{\lambda}^{(k)})
\end{equation}

It is found in practice that the quadratic term of the augmented Lagrangian of Equation (\ref{equ:AL2}) can dominate the loss function computation and lead to a local optimum. One solution to this problem is to perform the gradient calculation one step ahead to improve the convergence. This is referred to as the  {\em extra-gradient} method~\cite{Korpelevi1976An}.
\begin{equation}\label{equ:weightupdate}
  \begin{split}
  \bar{\boldsymbol{\Theta}} \leftarrow \boldsymbol\Theta^{(k)} - \eta_1\nabla L(\boldsymbol{\Theta})\\
  \boldsymbol{\Theta}^{(k+1)} \leftarrow \boldsymbol{\Theta}^{(k)} - \eta_2\nabla L(\bar{\boldsymbol{\Theta}})
\end{split}
\end{equation}

$\bar{\boldsymbol\Theta}$ here represents the temporary variable to store the intermediate backward parameters, and $\eta_1$ and $\eta_2$ are separate learning rates.

 \vspace{0.5em}
\noindent \textbf{2. Quantization variables update}
 \vspace{0.5em}

 The quantization discrete variables can be solved by minimizing the following:
 \begin{equation}
  \begin{split}
  &\min \limits_f ||\boldsymbol{\Theta}^{(k+1)}-f(\boldsymbol\Theta^{(k)})+\boldsymbol\lambda^{(k)}||_2^2 \\
  \Rightarrow&\min \limits_{\alpha, \textbf{Q}} ||\boldsymbol{\Theta}^{(k+1)}+\boldsymbol\lambda^{(k)}-\alpha\boldsymbol{V}^{(k)}||^2
\end{split}
\end{equation}

where $\boldsymbol{V}$ is calculated as

\begin{equation}
V_{i,\alpha}^{(k+1)} = \arg \min\limits_{q}|\Theta_i^{(k+1)} - \alpha q|
\end{equation}

The scaling factor $\alpha$ is then updated as 
\begin{equation}\label{equ:factor}
  \alpha^{(k+1)} = \frac{(\boldsymbol{\Theta}_i^{(k+1)}+\boldsymbol\lambda^{(k)})^\top\boldsymbol{V}_{i,\alpha}^{(k+1)}}{\boldsymbol{V}_{i,\alpha}^{(k+1)\top}\boldsymbol{V}_{i,\alpha}^{(k+1)}}
\end{equation}

$\alpha$ and $\boldsymbol{V}$ are updated interatively in an alternating way until convergence is reached.

 \vspace{0.5em}
\noindent\textbf{3. Error update}
 \vspace{0.5em}

The Lagrange multiplier variable $\boldsymbol\lambda$, now encoding the accumulated quantization errors computed at each iteration, is updated as
% he calculation of $\boldsymbol\lambda$ can be implemented from the updated weight and quantization table calculated from previous steps.
\begin{equation}\label{tml}
 \boldsymbol\lambda^{(k+1)} = \boldsymbol\lambda^{(k)} + \boldsymbol{\Theta}^{(k+1)}-f(\boldsymbol{\Theta}^{(k+1)})
\end{equation}

%The pseudo code of the entire ADMM based RNNLMs quantization algorithm is presented below in algorithm \ref{alg:Framwork}:

%\begin{algorithm}[htb]  
% \caption{ADMM based RNNLMs quantization }  
%  \label{alg:Framwork}  
%  \begin{algorithmic}[1]  
%    \Require  
%    Randomized full precision weight parameters $\boldsymbol\Theta_0$, initial quantization map table $f^0$ and scaling factor $\alpha_0$ set to be 1
    
%    \For{each iteration $k$}     
    
%   \State Compute the gradient of the loss function with respect to full precision weights using equation~(\ref{equ:gradient})~(\ref{equ:weightupdate}) with quantization table and scaling factor $\{\alpha^{(k)}\}$ fixed.
   
%   \State Given the updated full precision weight parameters $\boldsymbol\Theta^{(k+1)}$, update the quantization table without scaling factor and the scaling factors afterwards using equation~(\ref{equ:factor})
   
%   \State Update the multiplier (the quantization error variable) $\boldsymbol\lambda$ using equation~(\ref{tml})
    
%    \EndFor
     
%   \State  \Return  the best quantized values of the weight parameters
%  \end{algorithmic}  
%\end{algorithm}  

In all experiments of this paper the scaling factors $\alpha$ are initialized to one. The above ADMM quantization algorithm can be executed iteratively until convergence measured in terms of  validation data entropy is obtained. Alternatively, a fixed number of iterations, for example, 50, was used throughout this paper. The best performing model was then selected over all the intermediate quantizations obtained as each iteration. 

%The learning rates $\eta_1$ and $\eta_2$ in Equation (17) are set empirically as 0.1 and 1.0 respectively in all the experiments of this paper. And

\section{Experiments}
In this section, we evaluate the performance of quantized RNNLMs using the trade-off between the compression ratio and the perplexity (PPL) measure combined with word error rate (WER) obtained in automatic speech recognition (ASR) tasks. All the models are implemented using Python GPU computing library PyTorch~\cite{paszke2017automatic}. For all RNNLMs, the recurrent layer is set to be a single LSTM with 200 hidden nodes. 

In all models, parameters are updated in mini-batch mode (10 sentences per batch) using the standard stochastic gradient descent (SGD) algorithm with an initial learning rate of 1, optionally within the ADMM based quantization of section~\ref{sec:4}. In our experiments, all RNNLMs were also further interpolated with 4-gram LMs~\cite{emami2007empirical}~\cite{park2010improved}~\cite{le2012structured}. The weight of the 4-gram LM is determined using the EM algorithm on a validation set.

\subsection{Experiments on Penn Treebank Corpus}

We first analyze the performance of ADMM based quantization as well as binarized LSTM language model (BLLM~\cite{liu2018binarized}), BLLM without linear layer and the standard full precision RNNLM the  on the Penn Treebank (PTB) corpus, which consists of $10K$ vocabulary, $930K$ words for training, $74K$ words for development, and $82K$ words for testing. The PPL results are shown in Table 1.

\begin{table}[th]
  \caption{Performance and compression ratio of quantized LSTM RNNLMs on PTB corpus: full precision baseline with no quantization (STD), binarized model w/o  partially quantized linear layers (Binr+Lin or Bin) trained using 50 or 400 epochs with hidden size $H=200$, and ADMM quantized models with a layer, node or no tying of quantization tables of varying \#bits.}
  %Measurement on the perplexity of different quantization methods on PTB dataset, including Baseline (STD), Binarized LSTM LM~\cite{liu2018binarized} (Bin.), BLLM without linear layer (Bin.nl), ADMM based quantization (ADMM) on different quantization set (\textbf{Quan. Set} with scaling factor $\alpha$ omitted) and tying level. \textbf{Size} column gives the trained model size (MB) and compression ratio with respect to the baseline model and \textbf{PPL} means the perplexity of the trained model on test set.}
  \label{tab:ptb}
  \centering
  \resizebox{80mm}{27mm}{
  \begin{tabular}{l|c|c|c|c|c }
    \toprule
     \multirow{2}{*}
   { \textbf{RNNLMs}} & 
    \multirow{2}{*}{\textbf{Tying}} & 
    \multirow{2}{*}{\textbf{Quan. Set}} &\textbf{Model} &\textbf{Comp.}
    & \multirow{2}{*}{\textbf{PPL}} \\ 
         & & & \textbf{Size(MB)}  & \textbf{Ratio} & \\
    \midrule
    \textbf{STD}    \ \ \ \ \ \ \  10ep. & - & - & 16.53 & 1 & 114.4\\
    \hline
       \textbf{Bin} \qquad\  50ep.&  \multirow{4}{*}{-}  &  \multirow{4}{*}{$\{\pm \frac{1}{\sqrt{H}}\}$}  & 0.52 &  \multirow{2}{*}{31.8} & 131.6 \\
\cite{liu2018binarized}\qquad\ 400ep.  &  &  & 0.52 &  & 126.3 \\
\cline{1-1}\cline{4-6}
  \textbf{Bin+Lin}\ 50ep.  &   &  & 0.53 & \multirow{2}{*}{31.2} & 121.5\\
\cite{liu2018binarized}\qquad\ 400ep.  &   & & 0.53 & & 116.2\\
     \hline
    \multirow{9}{*}{\textbf{ADMM}\  \ \ (50ep.)}  & {NoTie}  & $\{1\}$ &  16.53&1 & 114.6\\ \cline{2-6}
                                      & \multirow{4}{*}{Node} & $\{\pm1\}$ & 0.65 &25.4 & 117.2 \\
                                      &  & $\{0\}\cup\{\pm1 \}$& 1.32 & 12.5& 119.3 \\
                                      &  & $\{\pm1, \pm2\}$ & 1.32 &12.5& 118.4\\
                                      &  & $\{\pm1, \pm2,\pm4\}$ &2.01& 8.2& \textbf{115.9}\\\cline{2-6}
                                      & \multirow{4}{*}{Layer} & $\{\pm1\}$ & 0.52 &31.8  & 121.8 \\
                                      &  & $\{0\}\cup\{\pm1 \}$& 1.06 &15.6 & 123.1 \\
                                      &  & $\{\pm1,\pm2\}$ &1.06 & 15.6 & 122.7\\
                                      &  & $\{\pm1, \pm2,\pm4\}$ &1.57 &10.5& 120.4\\
 
    \bottomrule
  \end{tabular}}
\end{table}

\begin{table*}[th]
  \caption{Perplexity (PPL) and word error rate (WER) performance and compression ratio of quantized LSTM RNNLMs on SWBD dataset: full precision baseline with no quantization (STD), binarized model w/o  partially quantized linear layers (Bin+Lin or Bin) trained using 50 or 400 epochs, and ADMM quantized models with a layer, node or no tying of quantization tables of varying \#bits. }
  \label{tab:swbd}
  \centering
\resizebox{170mm}{32mm}{
  \begin{tabular}{l r | c | c | c | c | c | c  c| c c}
    \toprule
  \multirow{2}{*}{\textbf{RNNLMs}} & & 
   \multirow{2}{*}{ \textbf{Tying}} & 
   \multirow{2}{*}{ \textbf{Quan. Set}}&\textbf{Model} &\textbf{Comp.}& \multirow{2}{*}{\textbf{PPL}} &\multicolumn{2}{c|} {\textbf{WER}} & \multicolumn{2}{c}{\textbf{WER}+4gram} \\
   & & & & \textbf{Size(MB)} & \textbf{ratio} & & swbd.&callhm. & swbd.& callhm \\
    \midrule
    \textbf{STD} & 8 ep.   & - & - & 48.11 & 1  & 89.3 & 11.4 & 23.9& 11.3 & 23.2 \\
  \hline
     \multirow{2}{*}{  \textbf{Bin.}} & 50 ep.&  \multirow{4}{*}{-} & \multirow{4}{*}{$\{\pm\frac{1}{\sqrt{H}}\}$} & \multirow{2}{*}{1.51} & \multirow{2}{*}{31.8} & 128.1 & 14.1 &29.0 & 13.5 &26.5  \\

      & 250ep.& & &  &  & 123.4 & 13.6 & 27.8 & 12.7 & 25.7  \\
      \cline{1-2}\cline{5-11}
   \multirow{2}{*}{ \textbf{Bin.+Lin.} } & 50ep. & &   & \multirow{2}{*}{1.52} & \multirow{2}{*}{31.6}  & 103.7 & 12.9& 25.8 & 11.9 & 24.9  \\

    & 250ep. &   & &  &   & 96.7 & 12.2 & 25.2 & 11.9 & 24.5 \\
    \hline
    \multirow{9}{*}{\textbf{ADMM}}  & \multirow{9}{*}{50ep.} & {NoTie}  & $\{1\}$ &  48.11 & 1 & 89.9 & 11.4 & 23.9 & 11.3 & 23.2 \\ \cline{3-11}
                                    &  & \multirow{4}{*}{Node} & $\{\pm1\}$ & 1.76  & 27.3& 98.2 &12.1 & 25.0 & 11.9 & 24.2 \\ 
                                    &  &  & $\{0\}\cup\{\pm1 \}$& 3.53 &13.6& 97.9 & 12.1 & 25.1& 11.8 & 24.4 \\
                                    &  &  & $\{\pm1,\pm2\}$ & 3.53 &13.6& 98.3  &12.2  & 25.1 & 11.9 & 24.4 \\
                                     & &  & $\{\pm1, \pm2, \pm4\}$ &5.30&9.1& 95.6 &11.9& 24.9 &\textbf{11.7} & \textbf{24.0} \\\cline{3-11}
                                     & & \multirow{4}{*}{Layer} & $\{\pm1\}$ & 1.51 &31.8 & 100.3 &12.2 & 25.3 & 11.9 & 24.6\\
                                     & &  & $\{0\}\cup\{\pm1 \}$& 3.10 & 15.5&101.1 & 12.8 & 25.3& 12.4 & 24.7 \\ 
                                     & &  & $\{\pm1,\pm2\}$ &3.10&15.5& 102.4 & 12.9& 25.5 & 12.5 & 24.9\\
                                      & &  & $\{\pm1, \pm2, \pm4\}$ &4.61 &10.4& 99.5  & 12.6 & 25.1& 12.3 & 24.4 \\
    
\bottomrule
\vspace{-2em}
  \end{tabular}}
\end{table*}

The performance of various quantized RNNLMs using binarization and ADMM optimization are presented in table~\ref{tab:ptb}. Full precision models are shown in table~\ref{tab:ptb} including the full precision baseline with no quantization (STD), binarized model w/o  partially quantized linear layers (Binr+Lin or Bin) trained using 50 or 400 epochs, and ADMM quantized models with a layer, node or no tying of quantization tables of varying \#bits.
% The  none-quantized full precision model is shown in line 1. The binarized RNNLM with/without the additional partial quantized linear layers and optionally with faster 50 epochs in training and longer training cycle of 400 epochs are shown in the table~\ref{tab:ptb} of line 2-5. 

There are several trends that can be found in table~\ref{tab:ptb}. First, the baseline binarization can obtain model compression factor up to approximately $31.8$ time. In order to achievw the best perplexity of $116.2$ (line5 in table~\ref{tab:ptb}), it requires both the additional partially quantized linear layers and 400 epochs ($103$s on GPU per epoch) to reach convergence in training. Also, there is a small perplexity increase by $1.8$ against the full precision baseline system. Second, the ADMM based quantization provides a much faster training, requiring 50 epochs (($193$s on GPU per epoch)) to converge for all ADMM quantized models, which is nearly 4 times faster in convergence over the Bin+lin binarized quantization baseline. Finally, the use of locally shared quantization tables in ADMM systems allow flexible adjustment in the trade-off between the model performance and compression ratio. To achieve the largest compression ratio of $31.8$, the layer level tied binarized ADMM based quantization system (line 11 in table~\ref{tab:ptb}) should be used. On the other hand, the best performance can be obtained using node level tying with $\{\pm1, \pm2, \pm4\}$ quantization table giving a perplexity score of 115.9 (line 10 in table~\ref{tab:ptb}).

%Compared with standard full precision RNNLM (\emph{STD}), by quantizing all the parameters into two fixed values ($\pm \frac{1}{\sqrt{H}}$), binarized LSTM language model without linear layers (\emph{Bin.nl}) can obtain $31.8$ compression rate and perplexity of 114.4. As suggested in~\cite{liu2018binarized}, two additional linear layers including binarized weight parameters and full precision bias are needed to mitigate performance degradation. This binarized model (\emph{Bin}) can perplexity of 116.2 with minuscule model size increasing. Our ADMM based quantization models are built on three different tying levels. For model without tying, each parameter has its own scaling factor, and the original full precision model can be reproduced under this setting. When the weight parameters of each node share the same scaling factor, we get the quantization model with node level tying. Up to 25.4 compression ratio can be obtained with acceptable performance degradation. More aggressive model compression ratio can be obtained with layer level tying that the parameters in each layer share the same scaling factor. With binary quantization set $\{\pm1\}$, the model produced in our method has almost the same model size as \emph{Bin.nl}, only additional three float scaling factor need to be stored. We can get the same compression ratio with better performance while maintaining the model structure. As can be observed, generally speaking, the more aggressive compression ratio we obtain, the more performance degradation the quantized model generates.

\subsection{Experiments on Conversational Telephone Speech}

%\begin{figure}[t]
 % \centering
%  \includegraphics[width=\linewidth]{total.png}
%  \caption{Schematic diagram of speech production.}
%  \label{fig:scatter}
%\end{figure}

To further evaluate the performance of proposed ADMM based quantization of RNNLMs for speech recognition, we also used the Switchboard (SWBD) conversational telephone dataset. The SWBD system has 300 hour of conversational telephone speech from Switchboard I for acoustic modeling, 3.6M words of acoustic transcription and 30k words lexicon for language modeling. The acoustic model is a minimum phone error (MPE) trained hybrid DNN, of which the details can be found in~\cite{liu2018limited}. The baseline 4-gram language model was used to generate the n-best lists. The Hub5'00 data set with Switchboard (swbd) and CallHome (callhm) subsets were used in evaluation.  Perplexity and n-best lists rescoring word error rate performance of various quantized RNNLMs using binarization and ADMM optimization are shown in table~\ref{tab:swbd}.
%N-best lists was used to generate the word error rate.
 %on the Switchboard (SWBD) English system consisting of 3.6M words for training and 50K words for development. 
 
\begin{figure}[htbp]
\setlength{\belowcaptionskip}{-1cm}
\begin{center}
\includegraphics[scale=0.26, trim = 0cm 1cm 0cm 1cm, clip]{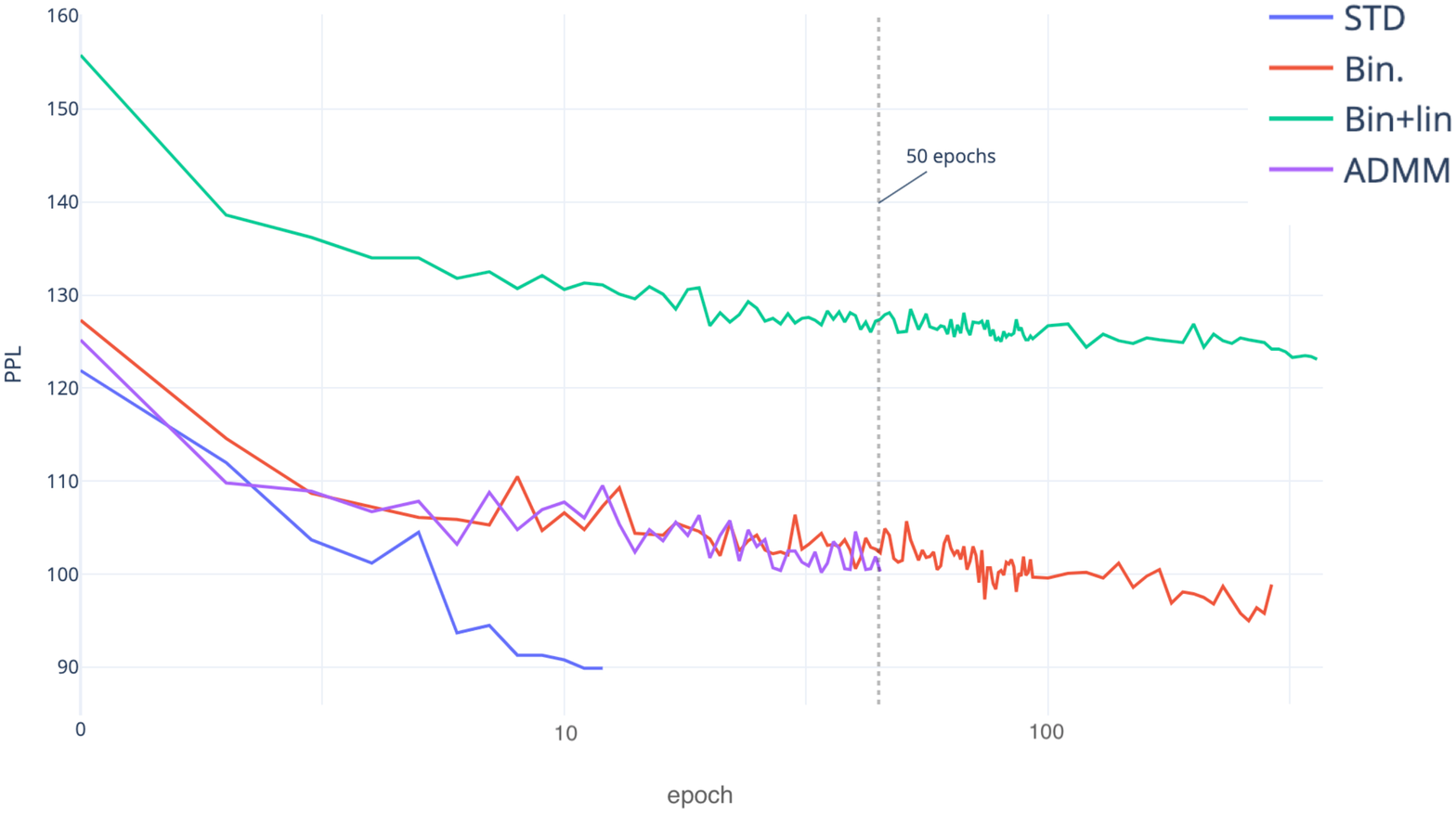}
\caption{Convergence speed comparison between the baseline full precision model, binarized RNNLM w/o additional linear layers and ADMM system with layer level tying and binary quantization set (line 11 in table~\ref{tab:swbd}) on SWBD}
\label{fig:per}
\end{center}
\end{figure}

A similar set of experiments as in table~\ref{tab:ptb} were then conducted on the SWBD data. Similar trends can be found. First, the ADMM based quantization convergences 3 times faster than the binarized RNNLM (Bin+lin) (line 5 in table~\ref{tab:swbd}). As can be seen from figure~\ref{fig:per}, our ADMM based quantization system only needs about 50 epochs (23min per epoch) to reach convergence, while the baseline binarized RNNLM (Bin and Bin+Lin) takes about as many as 250 epochs (10min per epoch). Second, the flexibility of ADMM based quantization in adjusting the trade-off between the compression ratio and model performance is clearly shown again in table 2. The largest compression ratio of $31.8$ was obtained using layer level tying and binary quantization set $\{\pm1\}$ (line 11 in table~\ref{tab:swbd}). The lowest word error rate 24.0 on callhm data and a compression ratio of $8.2$ was achieved by the node level tying quantization using $\{\pm1, \pm2, \pm4\}$ as the quantization set (line 10 in table~\ref{tab:swbd}). 

\section{Conclusions}

This paper investigates the use of alternating direction methods of multipliers (ADMM) based optimization method to directly train low-bit quantized RNNLMs from scratch. Experimental results conducted on multiple tasks suggest the proposed technique can achieve faster convergence than the baseline binarized RNNLMs quantization, while producing comparable model compression ratios. Future research will investigate the application of ADMM based quantization techniques to more advanced forms of neural language models and acoustic models for speech recognition.

\section{Acknowledgement}
This research is supported by Hong Kong Research Grants Council General Research Fund No.14200218 and Shun Hing Institute of Advanced Engineering Project No.MMT-p1-19.

\nocite{*}
\bibliographystyle{IEEEtran}
\bibliography{mybib}

% \begin{thebibliography}{9}
% \bibitem[1]{Davis80-COP}
%   S.\ B.\ Davis and P.\ Mermelstein,
%   ``Comparison of parametric representation for monosyllabic word recognition in continuously spoken sentences,''
%   \textit{IEEE Transactions on Acoustics, Speech and Signal Processing}, vol.~28, no.~4, pp.~357--366, 1980.
% \bibitem[2]{Rabiner89-ATO}
%   L.\ R.\ Rabiner,
%   ``A tutorial on hidden Markov models and selected applications in speech recognition,''
%   \textit{Proceedings of the IEEE}, vol.~77, no.~2, pp.~257-286, 1989.
% \bibitem[3]{Hastie09-TEO}
%   T.\ Hastie, R.\ Tibshirani, and J.\ Friedman,
%   \textit{The Elements of Statistical Learning -- Data Mining, Inference, and Prediction}.
%   New York: Springer, 2009.
% \bibitem[4]{YourName17-XXX}
%   F.\ Lastname1, F.\ Lastname2, and F.\ Lastname3,
%   ``Title of your INTERSPEECH 2019 publication,''
%   in \textit{Interspeech 2019 -- 20\textsuperscript{th} Annual Conference of the International Speech Communication Association, September 15-19, Graz, Austria, Proceedings, Proceedings}, 2019, pp.~100--104.
% \end{thebibliography}

\end{document}